\documentclass[10pt,twocolumn,letterpaper]{article}
\pdfoutput=1
\usepackage{cvpr}              

\def\cvprPaperID{10407} 
\def\confName{CVPR}
\def\confYear{2023}

\usepackage[utf8]{inputenc} 
\usepackage[T1]{fontenc}    
\usepackage[pagebackref,breaklinks,colorlinks]{hyperref}

\usepackage{url}            
\usepackage{booktabs}       
\usepackage{amsfonts}       
\usepackage{amsmath}
\usepackage{nicefrac}       
\usepackage{microtype}      
\usepackage{xcolor}         
\usepackage{caption}
\usepackage{placeins}
\usepackage{graphicx}
\usepackage{xspace}
\usepackage{subcaption}
\usepackage{array}
\usepackage[export]{adjustbox}
\usepackage[symbol]{footmisc}

\usepackage{color}
\definecolor{Blue9}{rgb}{0.098,0.3,0.9}

\newcommand{\ignore}[1]{}

\newif\ifdrafting
\draftingtrue 
\draftingfalse 
\ifdrafting
    \newcommand{\todo}[1]{{\leavevmode\color[rgb]{0,0,1}[TODO: #1]}}
    \newcommand{\ds}[1]{{\leavevmode\color[rgb]{1,0,0}[Deqing: #1]}}
    \newcommand{\ks}[1]{{\leavevmode\color[rgb]{0,0,1}[Kyle: #1]}}
    \newcommand{\cih}[1]{{\leavevmode\color[rgb]{0,0.4,0}[Charles: #1]}}
    \newcommand{\hc}{\color{Blue9}{hc: }}
    \newcommand{\jy}[1]{{\leavevmode\color[rgb]{1.0,0,1.0}[JY: #1]}}

\else
	\newcommand{\todo}[1]{}
    \newcommand{\ds}[1]{}
    \newcommand{\ks}[1]{}
    \newcommand{\cih}[1]{}
    \newcommand{\hc}[1]{}    
    \newcommand{\jy}[1]{}
\fi

\newcommand{\aroundeqn}{\vspace{-2pt}}
\newcommand{\beforepara}{\vspace{-6pt}}
\newcommand{\afterfigure}{\vspace{-8pt}}
\newcommand{\aftertable}{\vspace{-4pt}}

\newcommand{\modelname}{VQ3D\xspace}

\title{\modelname: Learning a 3D-Aware Generative Model on ImageNet}

\author{
  Kyle Sargent \\
  Stanford University\\
  \and
  Jing Yu Koh \\
  Carnegie Mellon University\\
  \and
  Han Zhang\\
  Google Research\\
  \and
  Huiwen Chang\\
  Google Research\\
  \and
  Charles Herrmann\\
  Google Research\\
  \and
  Pratul Srinivasan\\
  Google Research\\
  \and
  Jiajun Wu\\
  Stanford University\\
  \and
  Deqing Sun\\
  Google Research\\
}

\begin{document}

\maketitle

\begin{abstract}

Recent work has shown the possibility of training generative models of 3D content from 2D image collections on small datasets corresponding to a single object class, such as human faces, animal faces, or cars. However, these models struggle on larger, more complex datasets. To model diverse and unconstrained image collections such as ImageNet, we present \modelname, which introduces a NeRF-based decoder into a two-stage vector-quantized autoencoder. Our Stage 1 allows for the reconstruction of an input image and the ability to change the camera position around the image, and our Stage 2 allows for the generation of new 3D scenes. \modelname is capable of generating and reconstructing 3D-aware images from the 1000-class ImageNet dataset of 1.2 million training images. We achieve an ImageNet generation FID score of 16.8, compared to 69.8 for the next best baseline method. For video results, please see the project \href{https://kylesargent.github.io/vq3d}{webpage}.

\end{abstract}

\section{Introduction}

3D assets are an important part of popular media formats such as video games, movies, and computer graphics. Given that 3D content can be time-consuming to create by hand, leveraging machine learning techniques to automatically generate 3D content is an active area of research. While machine learning techniques benefit from training on large amounts of data, existing 3D datasets have noisy labels and are orders of magnitude smaller than those of 2D images. To get around the limitations of 3D datasets, recent work has shown the possibility of learning generative models of 3D scenes from images with limited or no 3D labels~\cite{giraffe:2020, eg3d:2021, gu2021stylenerf, hologan:2019}. 

These GAN-based approaches demonstrate the promise of learning 3D representations from 2D data. However, these methods require fine-tuning of prior pose distributions for individual models and datasets \cite{gu2021stylenerf, giraffe:2020, hologan:2019, pi-GAN:2020}, or the usage of ground truth pose data \cite{eg3d:2021}, 
and thereby typically operate on single-class datasets, e.g.,  human faces~\cite{karras2019style}, animal faces~\cite{afhq:2020}, or cars~\cite{compcars:2015}. 
In contrast, many 2D generative models, such as text-to-image generation models~\cite{parti, imagen:2022, ramesh2022hierarchical} and two-stage image models \cite{vqgan2020, yu2021vector} show impressive performance on very large and diverse image collections. The most recent state-of-the-art 2D models leverage diffusion or vector quantization rather than GANs to scale well to large datasets. This motivates us to pursue vector quantization as an alternative to GANs for learning 3D generative models.

\FloatBarrier
\begin{figure}[t!]
  \includegraphics[width=\columnwidth, right]{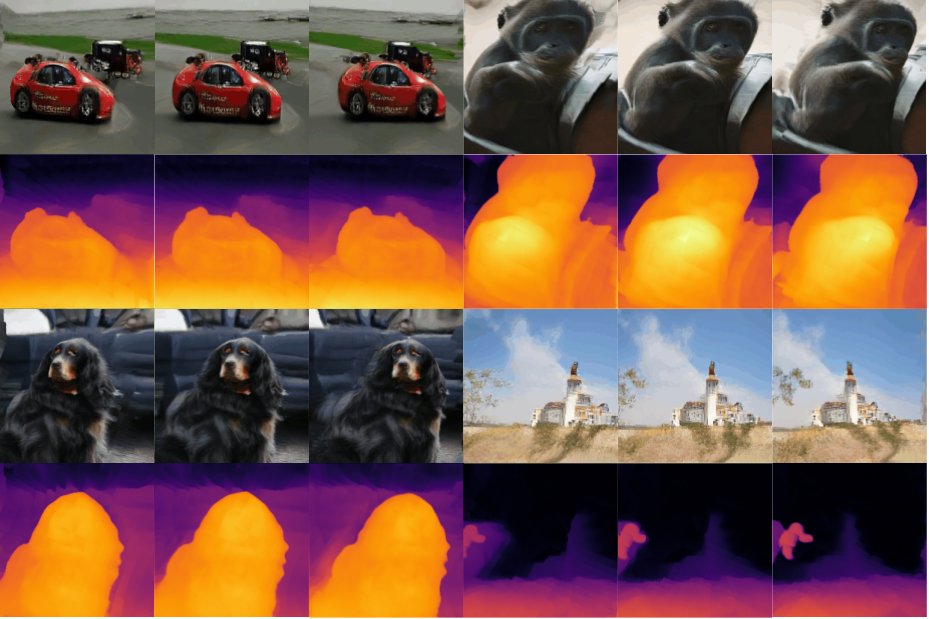}
  \caption{
  Fully generated 3D-aware images from our Stage 2 model on ImageNet. Please see supplemental materials for video results. } \afterfigure
\label{fig:teaser}
\end{figure}

In this paper, we propose \modelname, a strong 3D-aware generative model that can be learnt from large and diverse 2D image collections, such as ImageNet~\cite{imagenet:2009}.
To encourage stability and higher reconstruction quality, we forgo GAN-based~\cite{goodfellow2014generative} approaches~\cite{hologan:2019, pi-GAN:2020, eg3d:2021, giraffe:2020, gu2021stylenerf}, in favor of the 2-stage autoencoder formulation of VQGAN~\cite{vqgan2020} and ViT-VQGAN~\cite{ yu2021vector}. But, different from these 2D autoencoder models, we learn 3D geometry by introducing a conditional NeRF decoder and modified triplane representation which can handle unbounded scenes, and training with a novel loss formulation which encourages high-quality geometry and novel views.

Our formulation has three advantages, ensuring it to scale well to ImageNet. First, separating the training into two stages (reconstruction and generation) enables us to directly supervise the first stage training via a novel depth loss, using pseudo-GT depth. This is possible because in the first stage, as our conditional NeRF decoder learns to reconstruct the input, it also predicts the depth of each image. 

Second, we do not require hand-tuning of pose sampling distributions or ground-truth pose data, which are required by previous GAN-based approaches \cite{gu2021stylenerf, giraffe:2020, hologan:2019, pi-GAN:2020,eg3d:2021}. Our training objective simply enforces reconstruction from a canonical camera pose, and plausible novel views within a neighborhood of the canonical pose. While this objective regrettably rules out very large camera motion, it also eliminates the need for excessive tuning of the pose distribution for each dataset, and allows our model to work out-of-the-box for multiple object categories. Thus, our model uses identical pose sampling hyperparameters for each dataset. 

Finally, our two-stage formulation is simpler and more reliable than existing techniques for training 3D-aware generative models. Previous work~\cite{brock2018large, styleganxl:2022} has identified difficulties in scaling up GANs to large datasets (such as ImageNet). We verify that baseline 3D-aware GAN methods~\cite{eg3d:2021, gu2021stylenerf, pi-GAN:2020, giraffe:2020}, while working well on single-object datasets, fail to learn good generative models for ImageNet. Our formulation does not use progressive growing \cite{pi-GAN:2020, eg3d:2021}, a neural upsampler \cite{eg3d:2021, pi-GAN:2020, giraffe:2020}, pose conditioning \cite{eg3d:2021, epigraf}, or patch-wise discriminators \cite{Schwarz2020NEURIPS, epigraf}, but still learns meaningful 3D representations.  Compared to the best existing 3D-aware baseline, \modelname attains a 75.9\% relative improvement on FID scores for 3D-aware ImageNet images (69.8 for StyleNeRF~\cite{gu2021stylenerf} to 16.8 for \modelname).

In summary, we make the following three contributions:
\begin{itemize}
    \item We present a novel 
    3D-aware generative model that can be trained on large and diverse 2D image collections. Our model does not require tuning pose hyperparameters for each dataset or ground truth poses, and can leverage a pseudo-depth estimator during training. 
    \item We obtain state-of-the-art generation results on ImageNet and competitive results on CompCars, demonstrating that our 3D-aware generative model is capable of fitting a dataset at the scale and diversity of ImageNet. Our model significantly outperforms the next best baseline. 
    \item The Stage 1 of our model enables 3D-aware image editing and manipulation. One forward pass through our network converts a single RGB image into a manipulable NeRF, without relying on an expensive inversion optimization used in prior work~\cite{pi-GAN:2020, eg3d:2021}.
\end{itemize}

\section{Related Work}

\beforepara
\paragraph{3D-aware generative models.} 
Several recent papers tackle the task of modeling 3D-aware generation, primarily through the GAN framework~\cite{goodfellow2014generative}. HoloGAN~\cite{hologan:2019} learns perspective projection and rendering of 3D features, and applies 3D rigid-body transforms to generate new images from different poses. More recently, several papers use NeRF~\cite{mildenhall2020nerf} as the 3D backbone~\cite{giraffe:2020,pi-GAN:2020,gu2021stylenerf, xue2022giraffehd}, which allows the 3D scene to be defined as a 3D volume parameterized by an MLP. EG3D~\cite{eg3d:2021} proposes a hybrid triplane representation which scales well with resolution, and enables greater generation detail. Disentangled3D~\cite{tewari2022disentangled3d} learns a 3D-aware GAN from monocular images with disentangled geometry, appearance, and pose. Pix2NeRF~\cite{cai2022pix2nerf} proposes a method for unsupervised learning of neural representations with a shared pose prior, which enables rendering of novel views from a single input image. GRAF \cite{Schwarz2020NEURIPS} and EpiGRAF \cite{epigraf} train 3D GANs via patch-wise representations to save on the expense of volume rendering. GRAM~\cite{deng2022gram} proposes learning a set of implicit surfaces, shared for the training object category. At inference time, images are generated by accumulating the radiance along each ray using ray-surface intersections as samples.

\begin{figure*}
    \centering
	\newcommand{\Figwidth}{\linewidth}
	\includegraphics[width=\Figwidth]{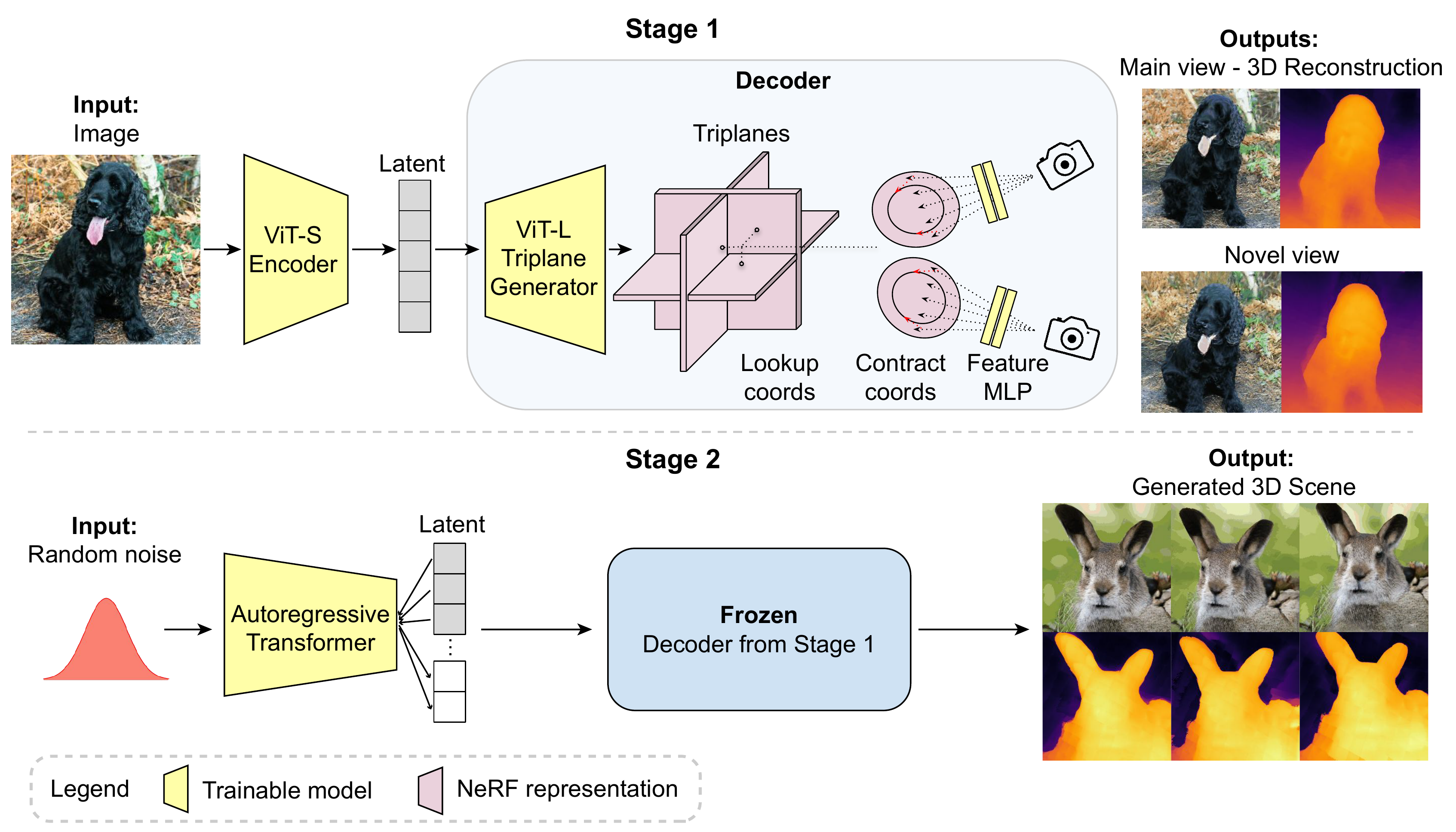}
    \caption{Diagram of our model architecture.}
    \label{fig:architecture}
\end{figure*}

\beforepara
\paragraph{Conditional NeRF and other 3D representations.} Recent work has focused on the appropriate way to condition NeRF to achieve maximum expressiveness. GIRAFFE~\cite{giraffe:2020} demonstrated success with the ``conditioning-by-concatenation'' approach~\cite{sitzmann2019scene}, in which the scene's latent codes are fed into the first layer of the NeRF MLP and not thereafter. Other work such as pi-GAN~\cite{pi-GAN:2020} transforms the latent code into a vector of frequencies and phase shifts for each layer of a SIREN~\cite{sitzmann2020implicit}. Other work has used hypernetworks~\cite{sitzmann2019scene, sitzmann2020metasdf} to parameterize 3D representations, and MetaSDF~\cite{sitzmann2020metasdf} showed that many forms of conditioning are special cases of the hypernetwork approach. Our model can be seen as a conditional NeRF. We show that our novel decoder architecture, consisting of a ViT-L \cite{dosovitskiy2020image} and contracted triplane representation, is powerful enough to encode and reconstruct all of ImageNet. Given a single image, we show that in a single forward pass and without any optimization, our model can create a NeRF of an input RGB image with reasonable reconstruction at the main view and plausible novel views.

\beforepara
\paragraph{Quantization models.} Image quantization is a powerful paradigm used in recent state-of-the-art generative models. In this setup, an image is encoded into a discrete latent representation~\cite{van2017neural}, which improves generation quality when paired with an autoregressive generative prior (most often a transformer~\cite{vaswani2017attention}). This has led to impressive results in image generation~\cite{esser2021taming,yu2021vector,lee2022autoregressive}, text-to-image generation~\cite{ramesh2021zero,ding2021cogview,parti}, and other tasks. Recent image quantization models improve reconstruction quality by introducing adversarial losses~\cite{esser2021taming}, using vision transformer encoders and decoders (ViT)~\cite{dosovitskiy2020image,yu2021vector} as both encoder and decoder, representing discrete codes as a stacked map~\cite{lee2022autoregressive}, and more. Such quantization architectures typically use powerful CNNs \cite{vqgan2020} or ViT \cite{yu2021vector} encoders and decoders;
ViT and CNN-based architectures show good performance reconstructing large image datasets; in this paper, we show that our NeRF-based decoder can also work well in the quantization framework. It has the capacity to encode and reconstruct a large and diverse dataset such as ImageNet, and also learns a discrete latent codebook that can be used to train a powerful fully generative Stage 2 model. 

\beforepara
\paragraph{Single-view 3D reconstruction and novel view synthesis.} Various approaches for 3D reconstruction or novel view synthesis in the context of generative or auto-encoder models have been proposed. Kato et. al \cite{kato2019learning} 
propose an adversarial training scheme using two discriminators for single-view 3D reconstruction. Their scheme, in which the main discriminator critiques real and reconstructed views, while an auxiliary discriminator distinguishes between the reconstructed input view and predicted novel views, inspires our use of two discriminators for similar reasons. However, their model cannot sample totally new scenes. 
More recently, uORF~\cite{yu2021unsupervised} uses NeRFs as 3D object representations to enable 3D scene decomposition. uORF represents a 3D scene as a composition of an object radiance field for each object, and a background radiance field for the remainder of the scene. This enables re-rendering and editing of 3D scenes from an input image. However, uORF also cannot sample new scenes, and moreover requires multi-view training datasets.

In the domain of novel scene generation, Generative Query Networks (GQN)~\cite{eslami2018neural} use CNNs to represent and generate scenes. GQNs can imagine and re-render scenes from novel viewpoints, but due to the usage of CNNs, do not explicitly embed 3D geometry or have any guarantees of scene consistency. NeRF-VAE~\cite{pmlr-v139-kosiorek21a} proposes an improved representation using a VAE which models multiple scenes. This enables efficient inference-time sampling of novel scenes, as well as re-rendering from multiple viewpoints. Unlike GQNs, which have no 3D prior, NeRF-VAE uses NeRF to achieve 3D consistency. However, it relies on multi-view training data. LOLNeRF \cite{lolnerf:2022} learns a generative model of 3D face images but requires a pretrained keypoint estimator and the auto-decoder formulation requires an optimization to be applied to examples outside its training set. By contrast, our method can be applied to single RGB images and requires only 2D training data and an off-the-shelf depth estimator for training.


\begin{figure*}
    \centering
	\newcommand{\Figwidth}{\linewidth}
	\includegraphics[width=\Figwidth]{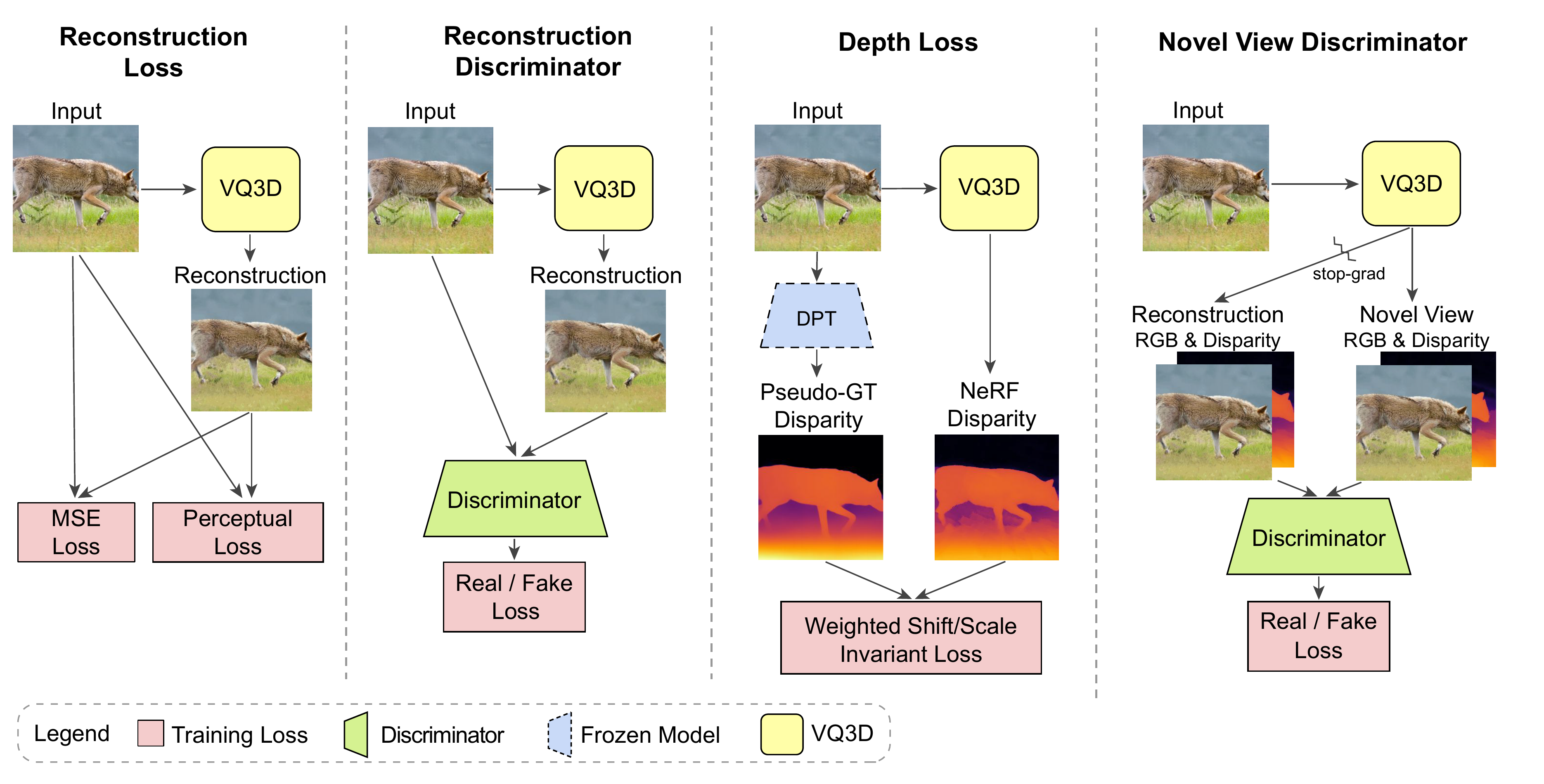} 
    \caption{Diagram of the key losses in Stage 1 optimization.
    } \afterfigure
    \label{fig:stage_1_optimization}
\end{figure*} 

\section{Model}

\beforepara
\subsection{Overview of VQ3D}
Our model is a vector-quantized autoencoder \cite{yu2021vector, vqgan2020}, which is trained in two stages.
Stage 1 of our model consists of an encoder and decoder. The encoder encodes RGB images into a learned latent codebook, and the decoder reconstructs them. A diagram of the inputs, outputs, and architecture of the first stage is given in the top of Figure \ref{fig:architecture}. The encoder of our first stage is a ViT similar to VIM~\cite{yu2021vector}, but the decoder is a conditional NeRF. The first stage is trained end-to-end by encoding and reconstructing RGB training images while minimizing reconstruction and adversarial losses. Because the decoder is a NeRF, we are able to supervise the NeRF geometry with an additional training loss using pseudo-GT disparity. We also render novel views of decoded images and critique them with an additional adversarial loss. A diagram of the key losses used in Stage 1 training is shown in Figure \ref{fig:stage_1_optimization}. After training, the first stage can be used to encode unseen single RGB images and then reconstruct them in 3D, which enables novel view synthesis, image editing and manipulations. 

Stage 2 is a generative autoregressive transformer which predicts sequences of latent tokens. A diagram of the inputs, outputs, and architecture is shown in the bottom of Figure \ref{fig:architecture}. The architectural and training details are generally the same as \cite{yu2021vector}. We train it on the sequences of latent codes produced by our Stage 1 encoder. After training, the autoregressive transformer can be used to generate totally new 3D images by first sampling a sequence of latent tokens and then applying our NeRF-based decoder. Importantly, our Stage 2 model inherits the properties optimized in Stage 1, so the fully generated images have high quality geometry and plausible novel views. 

\subsection{Training}

We now provide additional training details for the two stages of our model.

\beforepara
\paragraph{Stage 1.} The goal of the first stage is to learn a model which can compress image pixels into a sequence of discrete indices corresponding to a learnt latent codebook \cite{yu2021vector, vqgan2020}. Since we desire our model to be 3D-aware, we impose several additional criteria:
\begin{enumerate}
    \item \textbf{Good reconstruction from a canonical view.} On ImageNet, ground truth camera extrinsics are unknown and probably not even well-defined due to the presence of deformable and ambiguous object categories and scenes without salient objects. Therefore, we simply fix a single `canonical pose' for reconstruction, and our criterion is that our conditional NeRF-based autoencoder should successfully reconstruct the dataset from this view.
    \item \textbf{Reasonable novel views.} We expect that images decoded at novel views within a specified range of the canonical view will have similar quality to images decoded at the canonical view.
    \item \textbf{Correct geometry.} The geometry of the scene as represented by the NeRF should correspond to the unknown ground truth geometry of the RGB image up to scale and shift. 
\end{enumerate}

We enforce these criteria by introducing several auxiliary models and losses, summarized in Figure~\ref{fig:stage_1_optimization}. To enforce (1) good reconstruction at the canonical view, we train with a combination of the MSE, perceptual, and logit-laplace loss following \cite{yu2021vector}, the combination of which we term $\mathcal{L}_{\text{rec}}$.

To enforce (2) reasonable novel views, we leverage a main and auxilliary discriminator similar to \cite{kato2019learning}. The first discriminator distinguishes between real and reconstructed images at the canonical viewpoint, while the second distinguishes between reconstructed images at the canonical viewpoint and novel views. In this way, the model cannot allocate all its capacity to reconstructing images at the canonical viewpoint without also having high-quality novel views. As noted by \cite{kato2019learning}, the generator may slightly corrupt the main view in order to collaborate with the novel view branch to fool the discriminator; thus, we add a stop-grad between the main view and the novel view discriminator. Unlike \cite{pi-GAN:2020, gu2021stylenerf, eg3d:2021, giraffe:2020}, we find it unecessary to tune a separate distribution of novel views for each dataset, and instead sample novel views uniformly in a disc tangent to a sphere at the canonical camera pose. We use the non-saturating GAN objective $\mathcal{L}_{\text{gan}}$ ~\cite{goodfellow2014generative} for both discriminators. We additionally concatenate the predicted depth as input to the auxilliary discriminator to ensure the distribution of depths does not change depending on the camera viewpoint.

To enforce (3) correct geometry, we supervise the NeRF depth with pseudo-GT geometry at the main viewpoint. We employ the pretrained depth prediction transformer model DPT~\cite{ranftl2021dpt} which produces pseudo-GT disparity estimates for the images in our training datasets. Thus, our model is limited to some extent by the quality of the depth estimator chosen. \cite{ranftl2022midas} proposed a shift- and scale- invariant $l_2$ loss for training monocular depth estimation in which the shift and scale are determined by solving a closed-form least squares alignment with the GT depth. We propose a novel formulation of this shift- and scale- invariant loss adapted to the NeRF setting, in which we supervise the weight of every sample along each ray rather than the accumulated depth. For a given image, let $i \in \{1...N\}$ and $k \in \{1...L\}$ be indices which range over the image plane and ray samples respectively, let $D_{ik}$ be the pointwise disparities of the NeRF sample locations, let $W_{ik}$ be corresponding NeRF weights from volumetric rendering \cite{mildenhall2020nerf}, and let $d_i$ be the pseudo-GT depth from DPT. Then we define $s^*,t^*$ to be the closed-form solution of the weighted least squares problem
\aroundeqn
\begin{equation}
s^*,t^* = \arg\min \limits_{s,t} \frac{1}{N} \sum\limits_{i=1}^N \sum\limits_{k=1}^L W_{ik}(s D_{ik} + t - d_i)^2 
\end{equation}
\aroundeqn
And set our depth loss to be the weighted scale- and shift-invariant loss 
\aroundeqn
\begin{equation}
    \mathcal{L}_\text{depth} =  \frac{1}{N} \sum\limits_{i=1}^N \sum\limits_{d=1}^L W_{ik}(s^* D_{ik} + t^* - d_i)^2 
\end{equation}
\aroundeqn
Assuming the weight sum to 1 along each ray, this loss is minimized when the NeRF allocates 0 weight to all but one sample location along each ray, and the expectation with respect to the weights of the disparity is equal to the GT disparity map up to a scale and shift. In this way it functions similarly to the distortion loss proposed in \cite{barron2022mipnerf360} by penalizing weight distributions which are too spread out, but also encourages the weights to be concentrated near the correct geometry. Importantly, this formulation still allows for more than one surface along each ray and thus for occlusion and disocclusion, because the penalty is applied to the volumetric rendering weights and not the predicted density. We find this depth loss formulation to be critical for good performance. In particular, supervising the accumulated disparity rather than the pointwise disparities leads to poor performance, and we provide an ablation of this and other design choices in the supplementary material.  
We additionally introduce two penalties on the scale determined by this alignment:
\aroundeqn
\begin{equation}
    \mathcal{L}_\text{scale} = \lambda_{s1} \max(0, -s^*_\text{scale}) + \lambda_{s2} \max(s^*_\text{scale} - 1, 0)
\end{equation}
\aroundeqn
$\lambda_{\text{s1}}$ is the weight of a small penalty to prevent the sign of the disparity scale from flipping negative, which we found necessary unlike in \cite{ranftl2022midas}. $\lambda_{s2}$ weights a penalty preventing the disparity maps from becoming too flat, which encourages perceptually pleasing novel views.
We additionally include the same vector-quantization loss $\mathcal{L}_\text{vq}$ as \cite{yu2021vector}, and the distortion and interlevel losses of MipNeRF360\cite{barron2022mipnerf360}, given by $\mathcal{L}_\text{nerf}$. The loss for our autoencoder is thus:
\aroundeqn
\begin{equation}
\mathcal{L} = 
\mathcal{L}_{\text{rec}} + 
\mathcal{L}_{\text{gan}} + 
\mathcal{L}_{\text{depth}} + 
\mathcal{L}_{\text{scale}} + 
\mathcal{L}_{\text{vq}} + 
\mathcal{L}_{\text{nerf}}
\end{equation}
\aroundeqn

\vspace{-20pt}
\paragraph{Stage 2.}

The goal of Stage 2 is to learn an autoregressive model over the discrete encodings produced by the Stage 1 encoder, so that completely new 3D scenes can be generated. Our Stage 2 transformer and training details follow \cite{yu2021vector}. We verify experimentally that our fully generative Stage 2 model inherits the properties optimized in Stage 1; namely, 3D-consistent novel views and high quality geometry. We also apply top-$k$ and top-$p$ filtering similar to \cite{esser2021taming}.

\subsection{Architecture}
A full architecture diagram is shown in Figure~\ref{fig:architecture}. Similar to \cite{yu2021vector}, we leverage the powerful vision transformer~\cite{dosovitskiy2020image} architecture in both the encoder and decoder. Different from \cite{yu2021vector}, which is trained on 2D images, we utilize a novel decoder with 3D inductive bias to facilitate the learning of 3D representations. We now give an overview of the individual components of our architecture.

\beforepara
\paragraph{Encoder and triplane generator.}
For the encoder, we use a ViT-S model. For the decoder, we use a ViT-L model to decode the latent codes into 3 triplanes of size 512x512 with feature dimension 32. We find that the triplane construction stage of the decoder benefits from the increased capacity of the ViT-L model.

\beforepara
\paragraph{Contracted triplane representation \& NeRF MLP.} 
We must reconstruct and generate potentially unbounded ImageNet scenes, but we are motivated to leverage the powerful triplane representation~\cite{eg3d:2021}, Therefore, we propose an adapted triplane representation borrowing from both \cite{eg3d:2021} and \cite{barron2022mipnerf360}. We apply the contraction function of MipNeRF360 to bound coordinates within the triplanes before looking up their values, and use the linear-in-disparity sampling scheme with separate proposal and NeRF MLP. The MLPs convert interpolated triplane features to density and, in the case of the NeRF MLP, RGB color. Similar to \cite{eg3d:2021}, our MLPs are lightweight, with 2 layers and 32 hidden units each; unlike \cite{eg3d:2021}, we directly render RGB color rather than using a neural upsampler, as we found neural upsampling to be a source of myriad and confusing artifacts not fixable via dual discriminators~\cite{eg3d:2021} or consistency losses~\cite{gu2021stylenerf}.

\paragraph{Autoregressive transformer.} We train transformer~\cite{vaswani2017attention} to autoregressively predict the next image token. We follow the hyperparameters in the base model of VIM~\cite{yu2021vector}. For ImageNet, we train a conditional model, and for other datasets we train unconditional generative models. 

\section{Experiments}

\subsection{Main results} 

We study the performance of our method and the baseline methods on ImageNet. The ImageNet dataset~\cite{imagenet:2009} is a well-known classification benchmark which consists of 1.28M images of 1000 object classes. It is a standard benchmark for 2D image generation, for both conditional and unconditional generation. We compare against pi-GAN~\cite{pi-GAN:2020}, GIRAFFE~\cite{giraffe:2020}, EG3D~\cite{eg3d:2021}, and StyleNeRF~\cite{gu2021stylenerf}. We re-implented pi-GAN and GIRAFFE using our internal framework, and ran the provided code for EG3D and StyleNeRF. Since ImageNet does not have GT poses and pseudo-GT poses are not possible to compute, we disable generator and discriminator pose conditioning for EG3D and sample from a pre-defined pose distribution. We note that EG3D exhibits significant inter-run variance in ImageNet FID even for the same config, and provide more details in the supplementary material.

\begin{figure*}[!t]
  \centering
  \def\mywidth{0.3}
  \includegraphics[width=0.95\textwidth]{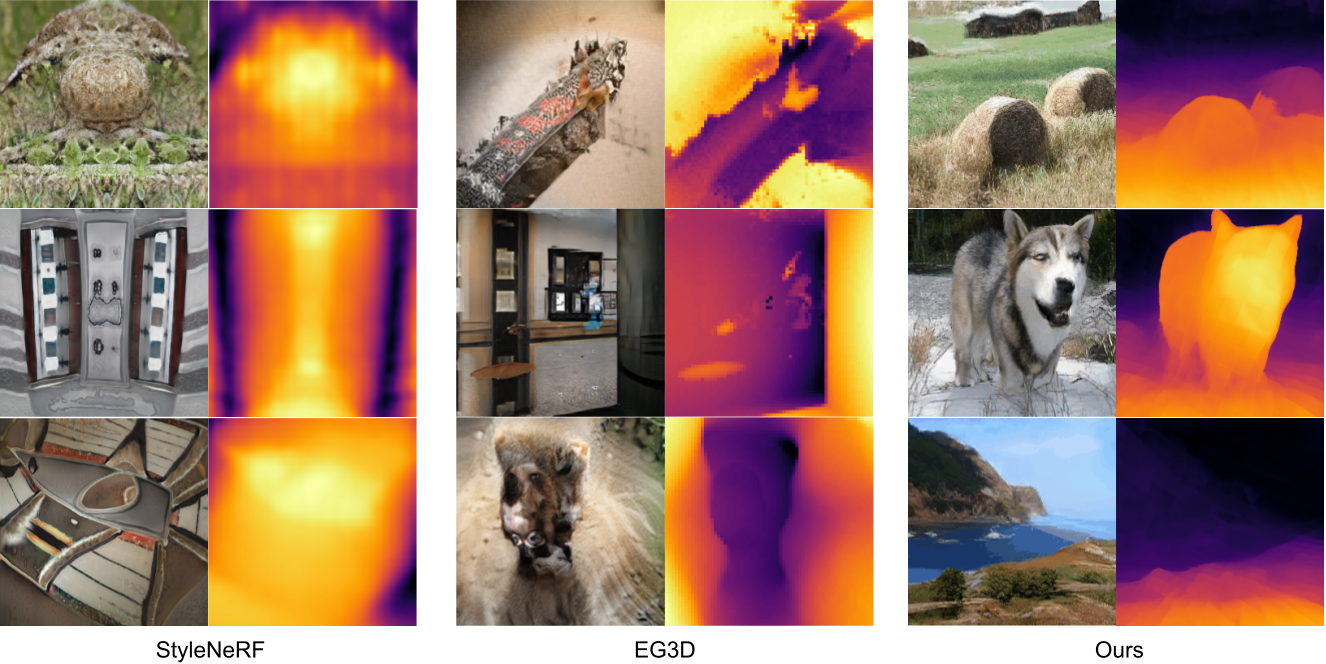}
  \vspace{-6pt}
  \caption{Generated samples and disparity from models trained on ImageNet. Ours model generates high-quality images and geometry. 
  }
\label{fig:qualitative_stage2}
\end{figure*}

Our main results for generation on ImageNet compared against the benchmarks are given in Table~\ref{tab:main_generation}. Notably, our FID score on ImageNet is the best by a wide margin.  We show generated examples from our method and the benchmarks in Figure~\ref{fig:qualitative_stage2} and note our method generates superior samples.

\begin{table}\centering\small
\begin{tabular}{lcc}
\toprule
\textbf{Generation} & \textbf{FID} $\downarrow$ \\
 \midrule
pi-GAN~\cite{pi-GAN:2020} & 97.8 \\
GIRAFFE~\cite{giraffe:2020} & 132.0 \\
StyleNeRF~\cite{gu2021stylenerf} & 69.8 \\
EG3D~\cite{eg3d:2021} & 82.2 \\
\midrule
\modelname (Ours) & \textbf{16.8} \\
\bottomrule
\end{tabular}
\caption{
FID scores of 3D generative models on ImageNet. We set a new state of the art on ImageNet with a more than fourfold improvement over the next best baseline. 
} \aftertable
\label{tab:main_generation}
\end{table} 

\begin{figure*}[!t]
  \centering
\includegraphics[width=\linewidth]{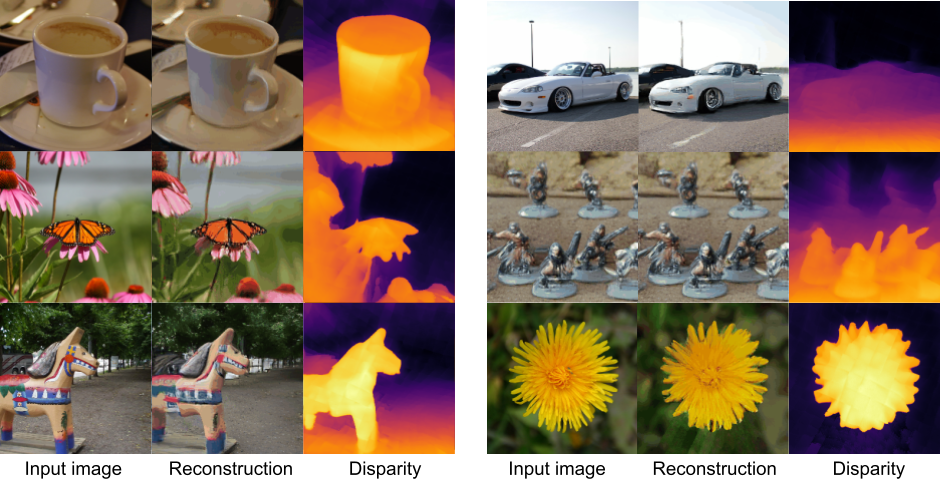}
\caption{Reconstructions and estimated disparity on single images by our conditional NeRF-based autoencoder. Though our model is trained on ImageNet and achieves comparable performance on unseen ImageNet images, we show OpenImages results for licensing reasons.} \afterfigure
\label{fig:qualitative_stage1}
\end{figure*}

\begin{figure}[!t]
  \includegraphics[width=\columnwidth]{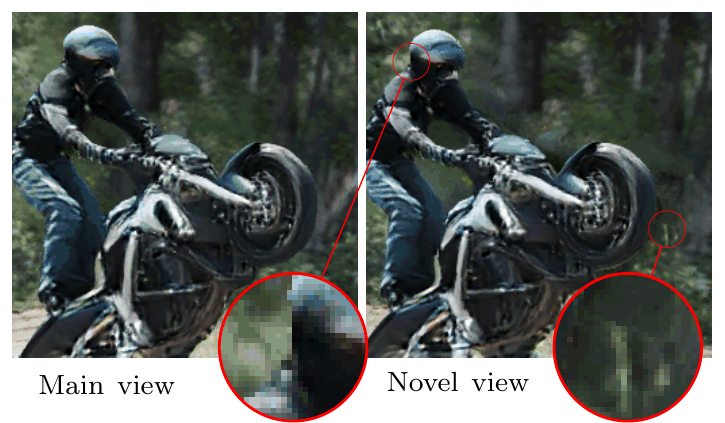}
  \caption{Example camera manipulations of a reconstructed scene. Our approach naturally handles sharp occlusions (left spyglass) and inpainting of disoccluded pixels (right spyglass) without supervision of novel views.
  } \afterfigure
\label{fig:novel_views}
\end{figure}

In addition to generating high quality scenes, Stage 1 of our method can also be used for single-view 3D reconstruction and manipulation. Figure \ref{fig:qualitative_stage1} shows single RGB images reconstructed by our Stage 1 with estimated geometry. Our network performs well at reconstruction and needs only a single forward pass to compute a NeRF for an input image, unlike prior work \cite{eg3d:2021, pi-GAN:2020} which requires an inversion optimization. Moreover, the reconstructed NeRFs can be manipulated, for instance to render novel views. We show examples of novel views in Figure \ref{fig:novel_views}. 
 
For our main results on ImageNet, we training for the longest possible time and use the most optimal top-p and top-k samping parameters. We conduct additional analysis experiments on the learning of geometry and model ablation, for which we use a consistent Stage 1 step, Stage 2 step across each study and do not using top-p or top-k sampling unless ablating it directly as in Table \ref{tab:topsampling}.

First, we study the learning of good geometry, both for our model and the baseline methods. One potential concern may be that the use of pseudo-GT depth limits the comparability of our technique with the baseline GAN methods. We address this concern by analyzing both the FID score and the depth accuracy metric used in \cite{liftsg:2021, eg3d:2021}. This metric is defined as the mean- and variance-normalized MSE between the NeRF depth and the predicted depth of the generated image. Table~\ref{tab:depth_ablation} gives the result for generative models with and without depth losses. Note that EG3D's FID without depth loss is different from Table \ref{tab:main_generation} due to the significant inter-run variance for EG3D's performance. For the GAN methods, we find that our pointwise disparity loss works poorly but the original scale- and shift- invariant MSE loss from \cite{ranftl2022midas} improves geometry. For our method, we show the Stage 2 performance with and without our novel pointwise weighted depth loss. While performance on the depth accuracy metric can improve when various depth losses are incorporated training, the effect on FID is negligible. In this way we see that incorporating pseudo-GT depth is unlikely to meaningfully improve the FID for the baseline methods without substantial changes. We were unable to design a depth loss which prevented flat depths for StyleNeRF. 

\begin{table}\centering\small
\begin{tabular}{lcc}
\toprule
\textbf{Generation} & \textbf{FID} $\downarrow$ & \textbf{Depth accuracy} $\downarrow$ \\
 \midrule
pi-GAN & 101.4 & 1.41  \\
GIRAFFE & 132.1 & 1.78  \\
EG3D & 109.3 & 1.66 \\
\modelname (Ours)  & 36.1 & 1.90  \\
\midrule
pi-GAN + depth loss & 97.8 & 0.88  \\
GIRAFFE + depth loss & 132.0 & 1.16  \\
EG3D + depth loss & 91.8 & 0.88 \\
\modelname (Ours) + depth loss & 31.7 & 0.16  \\
\bottomrule
\end{tabular}
\caption{Evaluation of depth losses on ImageNet. While adding depth losses can improve the quality of geometry, it will not lead to FID improvements significant enough to close the gap between our method and the baselines.
} \aftertable
\vspace{-2mm}
\label{tab:depth_ablation}
\end{table} 

Better geometry does not imply better FID. Additionally, learning geometry without a depth loss may be unreliable. For example, StyleNeRF \cite{gu2021stylenerf} found learning of geometry was unreliable without training tricks such as progressive growing. During our ImageNet experiments, we also observed that StyleNeRF is sensitive to hyperparameters, and can learn to produce flat depths. EG3D \cite{eg3d:2021} showed that removing GT poses as input to the discriminator is enough to cause the geometry to degenerate to a flat plane. 

We conduct ablations on our Stage 1 in Table \ref{tab:model_ablation} starting from our baseline architecture (row 1). Using a CNN encoder and decoder rather than ViT (row 2) is unstable and leads to divergence. Eliminating the GAN loss (row 3) or depth scale loss (row 4) leads to a higher learned disparity scale causing  perceptually flat novel views. Removing the GAN loss (row 3) also leads to artifacts in inpainting disoccluded pixels. Eliminating the NeRF loss (row 5) leads to worse depth accuracy and a very high disparity scale. Eliminating the depth loss (row 6) improves reconstruction FID, but causes the depths to collapse to a flat plane and leads to worse depth accuracy. A fully implicit representation instead of triplanes (row 7) gives very poor FID since we are forced to use a very small MLP due to the expense of volume rendering at 256x256.

\begin{table}\centering
\scriptsize
\begin{tabular}{lccc}
\textbf{Reconstruction} & \textbf{ImageNet FID}$\downarrow$ & \textbf{Depth Acc.}$\downarrow$ & \textbf{Disparity scale $\downarrow$} \\
\midrule
(1) VQ3D (Ours) & 11.2 & 0.18 & 1.00 \\
(2) CNN enc., dec. & (diverges) & - & - \\
(3) W/o $\mathcal{L}_{\textrm{gan}}$ & 10.6 & 0.22 & 1.27 \\
(4) W/o $\mathcal{L}_{\textrm{scale}}$ & 9.4 & 0.23 & 1.21 \\
(5) W/o $\mathcal{L}_{\textrm{nerf}}$ & 9.2 & 0.28 & 4.88 \\
(6) W/o $\mathcal{L}_{\textrm{depth}}$ & 4.0 & 1.91 & 0.61 \\
(7) W/o Triplanes & 273.5 & 1.00 & 2.15 \\
\end{tabular}

\caption{VQ3D ablation study. Removing components compromises the model capacity, 3D awareness, or novel view quality. \label{tab:model_ablation} }
\vspace{-5mm}
\end{table}

We analyze the performance of \modelname with top-$p$ and top-$k$ sampling in Table~\ref{tab:topsampling}, as \cite{vqgan2020} noted these sampling changes can give significant performance improvements analogous to truncation sampling for GANs \cite{brock2018large}. For \modelname, a top-$k$ of 1000 and top-$p$ of 1.0 gives the best FID results.

\begin{table}\centering\small
\begin{tabular}{llc}
\toprule
\textbf{top-$k$} & \textbf{top-$p$} & \textbf{FID $\downarrow$} \\
\midrule
1000 & 1.00 & 31.5 \\
2000 & 1.00 & 33.2 \\
3000 & 1.00 & 34.1 \\
4000 & 1.00 & 34.7 \\
8192 & 1.00 & 36.1 \\
8192 & 0.98 & 32.2 \\
8192 & 0.95 & 35.7 \\
\bottomrule
\end{tabular}
\caption{FID scores on ImageNet from sampling over top-$k$ and top-$p$ values. 8192 is the size of our full codebook. In general our model benefits from some restriction of the sampling process, though too restrictive top-p hurts performance.
} \aftertable
\label{tab:topsampling}
\end{table} 

\subsection{Other 3D benchmark datasets}

Two other prominent 3D-aware benchmark datasets are FFHQ \cite{karras2019style} and CompCars \cite{compcars:2015}. Due to the ethical and legal issues associated with manipulation and generative modeling of faces, we do not study FFHQ. On CompCars, our model is competitive with the state of the art (Table~\ref{tab:other_generation}). 

\begin{table}\centering\small
\begin{tabular}{lccc}
\toprule
\textbf{Generation} & \textbf{CompCars} \\
 \midrule
pi-GAN~\cite{pi-GAN:2020} & 16.9 \\
GIRAFFE~\cite{giraffe:2020} & $26^{\dagger}$ \\
StyleNeRF~\cite{gu2021stylenerf} & $8^{\dagger}$ \\
GIRAFFE HD\cite{xue2022giraffehd} & $7.2^{\dagger}$ \\
EG3D~\cite{eg3d:2021} & 32.2 \\
\midrule
\modelname (Ours) & 7.3 & \\
\bottomrule
\end{tabular} 
\caption{
FID scores of 3D generative models. $\dagger$ indicates numbers taken from the respective papers, we trained other models ourselves. Although baseline models use separate, tuned pose hyperparameters for each dataset, our identical, simple pose sampling scheme works well on both ImageNet and CompCars.} \aftertable
\label{tab:other_generation}
\end{table}

\section{Discussion and conclusion}

\beforepara
\paragraph{Limitations and ethical considerations.} Our work has several limitations. First, while some benchmark methods, e.g. \cite{giraffe:2020, gu2021stylenerf} have shown the ability to model 360-degree rotation of the generated scene when trained on specific single-class datasets like CompCars \cite{compcars:2015}, our auto-encoder based formulation makes large viewpoint manipulation difficult. While some baseline approaches such as \cite{eg3d:2021} and \cite{pi-GAN:2020} have demonstrated the ability to learn high-quality geometry in an unsupervised manner, our approach requires a pretrained depth network for the depth loss. However note that our model works on general object classes, while those focus only on small and single-class image collections and require tuning hyperparameters of the pose distribution.

We are committed to understanding and promoting positive societal impacts. Although we do not train a generative model on FFHQ and thus avoid many serious ethical considerations, ImageNet does contain some images of humans and human faces, and our model will likely inherit biases which are present in the dataset.


\beforepara
\paragraph{Conclusion.} We have presented \modelname, a framework for 3D-aware representation learning and generation.  \modelname sets a state-of-the-art by a wide margin on the large and diverse ImageNet dataset, relative to existing strong geometry-aware generative model baselines. We conduct extensive analysis and ablation and also show our model performs competitively on the more standard single class benchmark CompCars.
Our work shows that it could be a fruitful path to use large and diverse 2D image datasets to train 3D-aware generative models, thereby facilitating 3D content creation. 

\bibliography{refs}
\clearpage


Thanks for checking the supplementary materials. in which we provide additional details for the ease of replicating the results of our method. For video results, {\bf we encourage the reader to consult the project \href{https://kylesargent.github.io/vq3d}{webpage}}.

\section{Implementation details}

\subsection{NeRF Model}
We train and evaluate all models at 256x256 resolution, except pi-GAN \cite{pi-GAN:2020} which we train and evaluate at 128x128 following \cite{eg3d:2021}. 

We use a constant 49.13 degree field of view and pinhole camera model. We use a camera radius of 2.732 following \cite{giraffe:2020} and a canonical pose at $(-2.732, 0, 0)$. All views canonical and novel are looking at $(0, 0, 0)$ and have a constant camera up vector of $(0, 0, 1)$. We sample novel view camera locations uniformly in a disc in the YZ-plane centered at the canonical pose with radius $.4$. We use a near plane of $.7$ and far plane of $1e6$. We find that using the slightly large near plane of .7 was necessary in order to avoid a failure mode where all the content was clustered very close to the camera leading to poor novel views; we hope to eliminate this failure mode in future work.

We perform volume rendering at the full 256x256 resolution using the importance sampling scheme of \cite{barron2022mipnerf360}. We have a separate proposal and NeRF MLP and render in two stages, the first stage using the proposal MLP to evaluate a wide range of sample locations, and the second stage using the NeRF MLP queried at locations determined by importance sampling of the weights and locations from the first stage. During training, we add a stop-grad between the proposal and NeRF MLP like \cite{barron2022mipnerf360} and supervise the Proposal MLP with the interlevel loss. Our NeRF MLP is not view dependent and the only input it receives is triplane features which are determined by looking up the contracted 3D points of the sample locations. We apply a fixed orthonormal transformation to all points before triplane lookup because our canonical pose is axis-aligned, so we desire that our triplanes are not axis-aligned to avoid artifacts.

We evaluate 32 samples along each ray for each sampling stage. Thus, rendering a full 256x256 RGB image takes 256x256x64 triplane lookups and MLP evaluations. We use the same number of ray samples, 32, for training, FID evaluation, and rendering videos. 

\subsection{Setup and hyperparameters}
We train with the Adam optimizer \cite{kingma2014adam} with $\beta_1 = .9, \beta_2 = .99$, and cosine learning rate schedule with 50K warmup steps, similar to \cite{yu2021vector}, with an initial autoencoder LR of 0 and max LR of 1e-4. We use codebook size 8192 and $l_2$-normalized, factorized codebook with embedding dimension 8. 

Different from \cite{yu2021vector}, we do not use weight decay, and our discriminator LR is scaled down from the autoencoder LR by .5 so that the discriminator does not overpower the autoencoder, which was an issue especially in early training.

Due to the many losses in our Stage 1 training, we outline their weights in Table \ref{tab:losses} and reference the original implementation if they are not losses designed by us.

\begin{table}\centering
\begin{tabular}{lr}
\toprule
\textbf{Loss} & Weight \\
\midrule
$l_2$ \cite{yu2021vector} &             1 \\
Perceptual \cite{yu2021vector}  &       1e-1 \\
Logit-laplace \cite{yu2021vector} &     1e-1 \\
Discriminator \cite{yu2021vector}  &             1e-1 \\
Novel discriminator &                            1e-1 \\
Quantizer \cite{yu2021vector}  &        1 \\
Weighted pointwise depth ($\lambda_{\text{depth}}$) &              1e1 \\
Negative depth scale penalty ($\lambda_{s1}$) & 1 \\
Large depth scale penalty ($\lambda_{s2}$) & 1e-3 \\
Interlevel \cite{barron2022mipnerf360} & 1 \\
Distortion \cite{barron2022mipnerf360} & 2.5e-1 \\
\bottomrule
\end{tabular}
\caption{Weights of various losses used in Stage 1 training of our autoencoder.
\label{tab:losses}}
\end{table} 

\subsection{Discriminators}
We use StyleGAN \cite{karras2019style} discriminators for both the main and novel view discriminator. They are identical except that the novel view discriminator accepts 4-channel RGBD images, and the main view discriminator accepts 3-channel RGB images.

\subsection{Timing and throughput} We train our main model on ImageNet for 180K steps in Stage 1, and 140K steps in Stage 2. On a single V100, our Stage 1 model renders 8.7 img/s. We train with a Stage 1 batch size of 128 and Stage 2 batch size of 512. For each batch in Stage 1, we render 256 images; 128 to reconstruct the full batch at the canonical view, and an additional 128 novel views to be critiqued by the novel view discriminator. Though this is expensive, our volume rendering stage is made cheaper even than \cite{eg3d:2021} by using 32 instead of 64 hidden units for the feature MLPs and using 32 instead of 48 samples per ray. We leverage gradient accumulation in Stage 2 training in order to train with 512 batch size.

\subsection{Evaluation}
As is standard \cite{yu2021vector}, we compute Stage 1 metrics (reconstruction) over the ImageNet validation set and Stage 2 metrics (generation) over real samples from the train set and generated samples. We use 50K samples to evaluate FID for all methods. We sample views for Stage 2 FID computation uniformly in a disc of radius .2 tangent to the sphere at the canonical pose.

We use the Depth Accuracy metric used in \cite{liftsg:2021, eg3d:2021}, but differently we don't mask out any invalid regions because our monocular depth estimator DPT \cite{ranftl2021dpt} predicts a dense depth map over the input and every pixel is assumed to be valid. We also use disparity instead of depth because we model much larger scenes than either \cite{liftsg:2021} or \cite{eg3d:2021}.

We experiment with classifier guidance but find it gives only a small performance boost, and so investigating model improvements was more worthwhile to improve the FID than tweaking classifier guidance settings.

\section{Additional experiments}



Although the strongest baselines, EG3D \cite{eg3d:2021} and StyleNeRF \cite{gu2021stylenerf}, perform poorly on ImageNet, they may need to be tuned to perform well on this new dataset. To verify that the limitation of the baseline methods is fundamental, we extensively tune both on Imagenet for a range of hyperparameters in Tables \ref{tab:eg3d_hparam_tune} and \ref{tab:stylenerf_hparam_tune}. We see the baselines do not achieve good performance for a range of hyperparameter settings. Additionally, we observe that EG3D has significant inter-run variance in terms of FID on ImageNet, even when rerunning the same configuration, which may indicate instability for large datasets such as ImageNet. When running the same config multiple times, we report the best value achieved among all runs.

\begin{table}\centering
\scriptsize
\begin{tabular}{lccc}
\textbf{EG3D Tuning} & \textbf{Sweep} & \textbf{ImageNet FID} \\
 \midrule
R1 gamma & \{.3, .6\} & \{\textbf{82}, 99\} \\
Density reg. & \{.125, .25, .5\} & \{91, \textbf{82}, 96\} \\
Disc. LR (1e-3) & \{.5, 1, 2, 4\} & \{122, \textbf{82}, 116, 113\} \\
Gen. LR (1e-3)& \{.625, 1.25, 2.5, 5\} & \{111, \textbf{82}, 106, 136\} \\
\end{tabular}
\caption{
Hyperparameter tuning of EG3D on ImageNet. 
\label{tab:eg3d_hparam_tune}
}
\end{table}

\begin{table}\centering
\scriptsize
\begin{tabular}{lccc}
\textbf{StyleNeRF Tuning} & \textbf{Sweep} & \textbf{ImageNet FID} \\
 \midrule
R1 gamma & \{.15, .3, .6\} & \{75, \textbf{73}, 74\} \\
Disc. LR (1e-3) &\{.625, 1.25, 2.5, 5\} & \{96, 87, 73, \textbf{69}\}\\
Gen. LR (1e-3)& \{.625, 1.25, 2.5, 5\} & \{78, 74, \textbf{73}, 107\} \\
\end{tabular}
\caption{
Hyperparameter tuning of StyleNeRF on ImageNet.\label{tab:stylenerf_hparam_tune}}
\end{table} 

\section{Additional samples}

\begin{figure*}
\centering
\begin{tabular}{cc}
\includegraphics[width=.87\columnwidth]{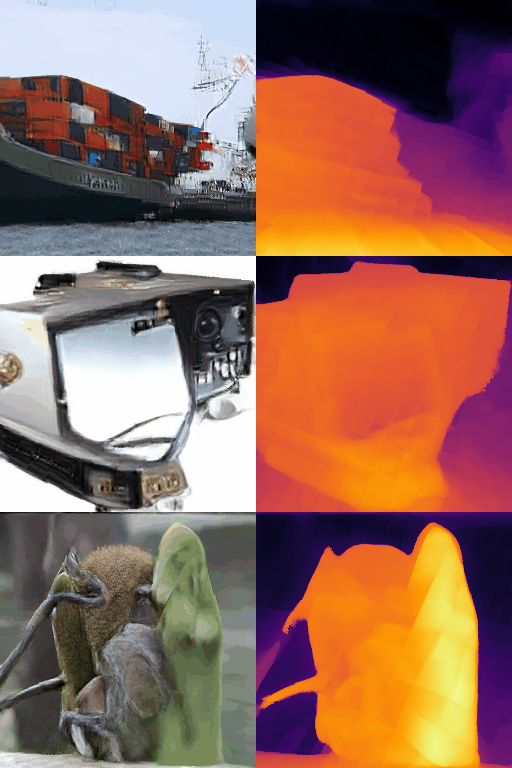} & \includegraphics[width=.87\columnwidth]{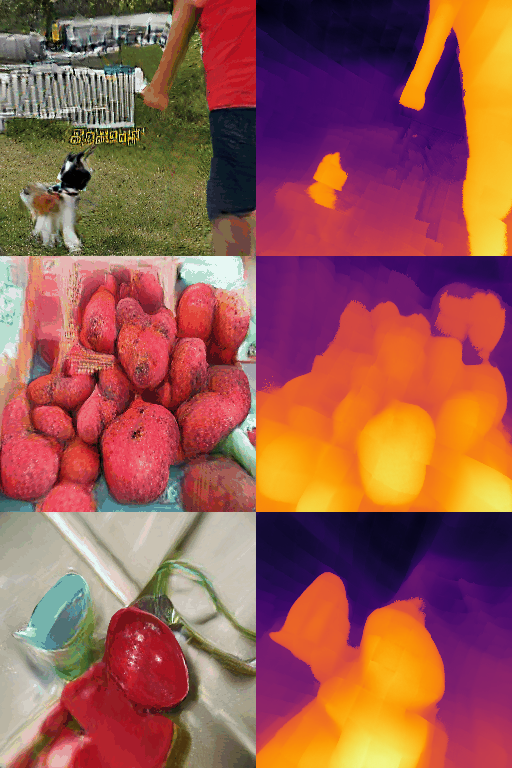} 
\vspace{7pt}
\\
\includegraphics[width=.87\columnwidth]{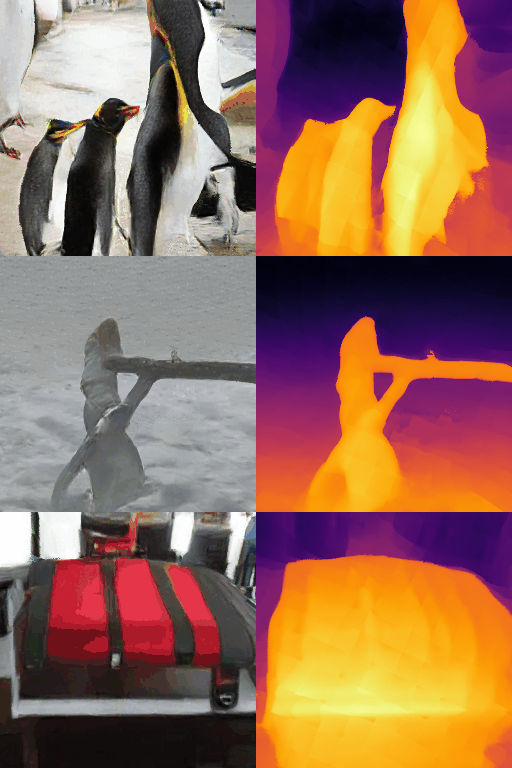} & \includegraphics[width=.87\columnwidth]{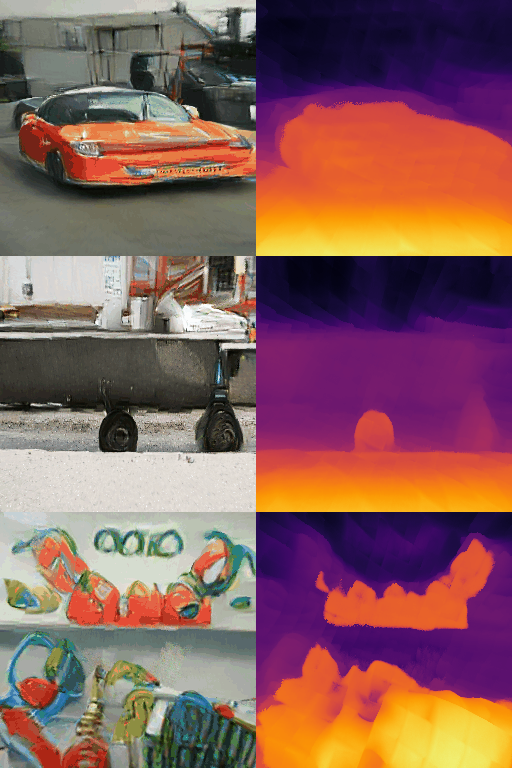} \\
\end{tabular}
\caption{Uncurated fully generated samples from our Stage 2 model. \label{fig:samples}}
\end{figure*}

\begin{figure*}
\centering
\begin{tabular}{cc}
\includegraphics[width=.87\columnwidth]{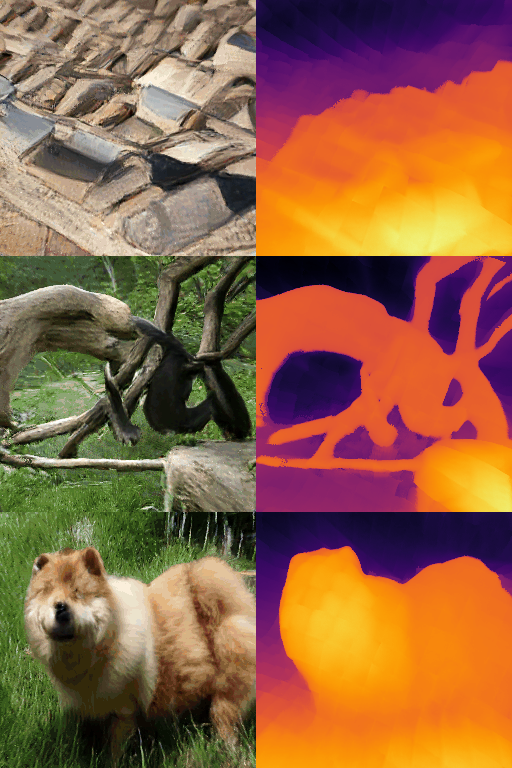} & \includegraphics[width=.87\columnwidth]{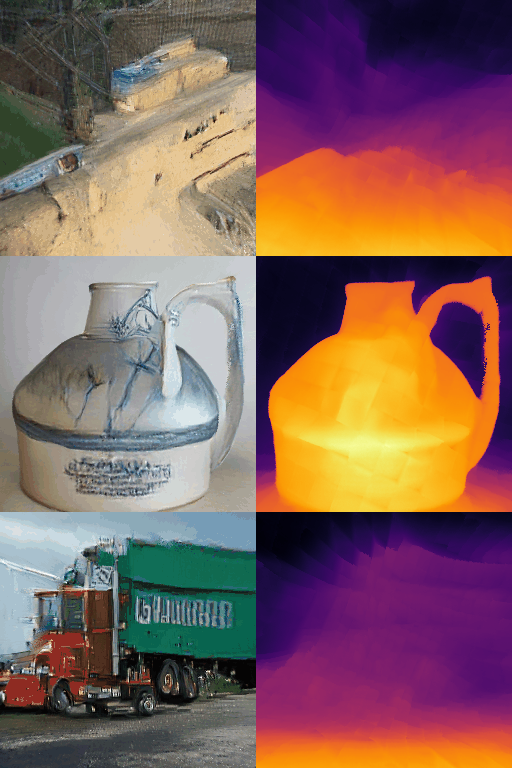}
\vspace{7pt}
\\
\includegraphics[width=.87\columnwidth]{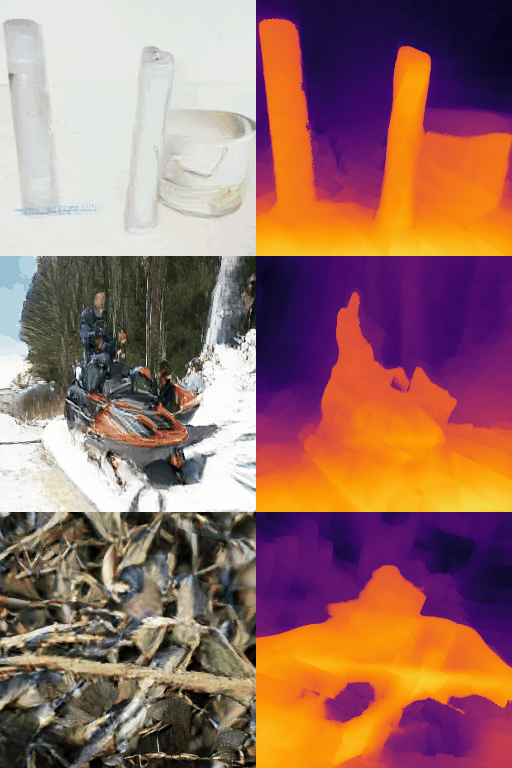} & \includegraphics[width=.87\columnwidth]{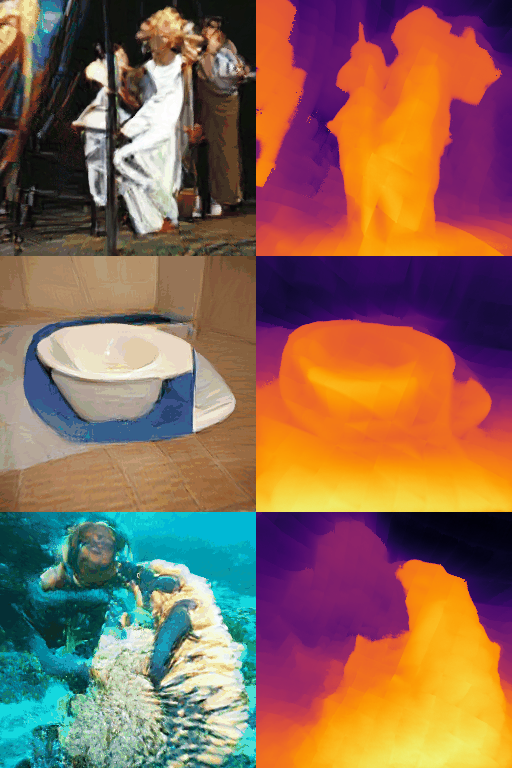} \\
\end{tabular}
\caption{More uncurated fully generated samples from our Stage 2 model. \label{fig:samples_2}}
\end{figure*}

We show additional uncurated generated samples with geometry in Figure \ref{fig:samples} and Figure \ref{fig:samples_2}.

\end{document}